# Vision-Based System Identification of a Quadrotor

Selim Ahmet IZ and Mustafa UNEL
*Faculty of Engineering and Natural Sciences*
Sabanci University
Istanbul, Turkey
{izselim, munel}@sabanciuniv.edu

*Abstract*— This paper explores the application of vision-based system identification techniques in quadrotor modeling and control. Through experiments and analysis, we address the complexities and limitations of quadrotor modeling, particularly in relation to thrust and drag coefficients. Grey-box modeling is employed to mitigate uncertainties, and the effectiveness of an onboard vision system is evaluated. An LQR controller is designed based on a system identification model using data from the onboard vision system. The results demonstrate consistent performance between the models, validating the efficacy of vision-based system identification. This study highlights the potential of vision-based techniques in enhancing quadrotor modeling and control, contributing to improved performance and operational capabilities. Our findings provide insights into the usability and consistency of these techniques, paving the way for future research in quadrotor performance enhancement, fault detection, and decision-making processes.

## I. INTRODUCTION

Quadrotor modeling is inherently challenging due to its highly nonlinear structure. The complex dynamics involved, such as aerodynamic interactions, variable thrust-to-weight ratio, and nontrivial coupling effects, make it difficult to accurately represent the system mathematically. Conventional modeling techniques often struggle to capture the intricacies of quadrotors, leading to suboptimal control performance. To address these challenges, advanced indoor motion tracking systems have emerged as a preferred approach for quadrotor modeling [1][2]. These systems utilize vision-based techniques, such as cameras and markers, to track the quadrotor's motion with high precision and accuracy [3][4]. By leveraging visual information, these systems can provide valuable data for system identification, enabling more accurate and reliable modeling of the quadrotor dynamics. The utilization of vision-based techniques opens up new possibilities for enhancing quadrotor modeling and control strategies.

System identification characterizes dynamic behavior of nonlinear systems like quadrotors using observed input-output data. It estimates mathematical models that describe the relationships between system inputs and outputs. The main purpose is to gain insights into the underlying dynamics and unknown parameters of the system which provide better understanding of system behavior [5], design effective control strategies [6], and making predictions about its future performance [7].

In recent years, advancements in quadrotor modeling through system identification techniques have emerged. State-of-the-art studies employ advanced algorithms and data-driven approaches to accurately capture the quadrotor's dynamics. Utilizing inputs and outputs like motor speeds, control inputs, and position measurements, these studies estimate high-fidelity models that represent the quadrotor's behavior. These studies have explored various approaches, including parameter estimation [8][9], model structure selection [10], and nonlinear system identification methods [11], to capture the intricate dynamics of quadrotors. Most quadrotor modeling studies using system identification techniques have focused on advanced motion-tracking systems rather than vision-based approaches. Motion tracking systems like GPS or motion capture provide accurate data for system identification, enabling precise modeling of quadrotor dynamics. While there have been advancements in vision-based system identification techniques, studies specifically targeting quadrotor modeling using vision-based methods are limited. Therefore, further research is needed to explore the potential of vision-based system identification for quadrotor modeling.

In this article, our goal is to explore the application of vision-based system identification techniques in quadrotor modeling. Initially, we consider a known white-box model, but due to uncertainties in thrust and drag coefficients under different environmental conditions, we employ grey-box modeling. Carefully selecting input and output parameters, we highlight the significance of these coefficients in quadrotor modeling. We choose a suitable controller for performance comparison and conduct experiments to identify the black-box system model using data from the onboard camera and marker system. The results of these models, along with experimental flight data, are compared in a MATLAB-Simulink environment. Our contributions are summarized as follows:

- Proposed a method to evaluate onboard vision system usability using system identification techniques.
- Detailed the process of obtaining the quadrotor's state space model through grey-box identification and RLS parameter estimation for LQR controller design.
- Distinguished between control input and rotor angular velocities in quadrotor system modeling.
- Highlighted the impact of thrust and drag coefficients on quadrotor modeling.

The following parts of this paper is organized as Section II where the methodology of the study shared by following quadrotor modeling, controller design, and system identification process titles, the section III where the obtained experimental results are compared and discussed, and finally section IV where the proposed approach and highlighted ideas are concluded.

## II. METHODOLOGY

The black-box quadrotor model's consistency can be assessed by comparing its performance to that of the grey-box system model under the control of an identical controller. This comparison investigates the usability and accuracy of the onboard sensing system used in the black-box modeling process.

### A. Quadrotor Modeling

The quadrotor system's underactuated and nonlinear nature poses challenges. Despite having only four control inputs tied to motor speeds, quadrotors exhibit six degrees of freedom involving three translational and three rotational axes. This complexity arises when the quadrotor moves solely on the vertical axis, affecting attitude states. However, controlling all six degrees of freedom using just four inputs (thrust, roll angle, thrust angle, and rotor speed difference) is not feasible. Additional control inputs are needed to command the remaining axes for simultaneous control of all six degrees of freedom, including vertical movement, roll, pitch, and yaw.

Assumptions are made to simplify the equations of motion and address external and structural disturbances impacting system dynamics [12][13]. Order reduction techniques can be used to handle the complexity and nonlinearity of the system's 12 control states, including positions, orientation, rotational speeds, and linear accelerations. Assumptions regarding the rigid and symmetrical structure of the quadrotor and neglecting blade flexibility are necessary in formulating the systemic equations for quadrotor modeling [14]. The thrust and drag of each motor should be considered proportional to the square of the motor velocity [15]. Simplifications in the modeling process can be achieved by disregarding external disturbances.

After making these assumptions, the equation of motion for quadrotors can be derived through sub-analyses, including translational kinematics, rotational kinematics, and electric-rotor analysis. Since quadrotors are electro-mechanical systems with four electric rotors, these analyses incorporate electro-mechanical considerations. In translational kinematics [16], a rotation matrix should be defined to transform variables between global and local coordinate axes:

$$R_b^G = \begin{bmatrix} c\psi c\theta & c\psi s\phi s\theta - c\phi s\psi & c\phi c\psi s\theta + s\phi s\psi \\ c\theta s\psi & s\psi s\phi s\theta + c\phi c\psi & s\psi c\phi s\theta - s\phi c\psi \\ -s\phi & s\phi c\theta & c\theta c\phi \end{bmatrix} \quad (1)$$

where $\phi$, $\theta$, $\psi$ are roll, pitch and yaw angle.

In addition, the rotational kinematics [17] play a crucial role in the quadrotor modeling process. The kinematics explain the relationship between body frame angular rates and Euler angles defined within coordinate frames. Rotational speeds (p, q, r) are calculated using this rotational kinematics. Thus, addressing rotational kinematics is as crucial as considering translational kinematics for accurate quadrotor system modeling.

$$\begin{bmatrix} \dot{\phi} \\ \dot{\theta} \\ \dot{\psi} \end{bmatrix} = \begin{bmatrix} 1 & s\phi t\theta & c\phi t\theta \\ 0 & c\psi & s\phi \\ 0 & s\phi/c\theta & c\phi/c\theta \end{bmatrix} \begin{bmatrix} p \\ q \\ r \end{bmatrix} \quad (2)$$

Once the kinematics computations are complete, the next crucial step is to consider the motion equations governing the behavior of the motors in the quadrotor system. As quadrotors fall under the category of electro-mechanical systems [18], it is important to analyze the power, torque, and velocity equations in terms of current and voltage. Various factors come into play during motor operation, including motor torque ($\tau$), motor power ($P_m$), torque constant ($K_\tau$), input current ($I$), current without load ($I_0$), voltage ($V$), motor resistance ($R_m$), back EMF coefficient ($K_v$), and angular velocity ($\omega$) of the motors. Accounting for these factors is essential in accurately modeling the dynamics of the quadrotor's motors.

$$P_m = \frac{(\tau + K_\tau I_0)(R_t I_0 R_m + \tau R_m + K_\tau K_v \omega)}{K_\tau^2} \quad (3)$$

In addition to the mentioned computations, another crucial assumption must be made regarding the quadrotor's behavior. The quadrotor's stability is compromised and its power consumption and motor performance are directly affected when the required current changes during movements other than hovering. Therefore, these calculations should be performed considering an optimal usage style, particularly the hover position. Moreover, it is important to incorporate the physical specifications of the quadrotor and environmental effects [12][13], such as air density, into the motion equations. At this stage, the drag force ($F_d$), drag coefficients ($K_d$) and thrust coefficient ($K_T$) play a significant role and must be accurately determined to obtain a proper model. These coefficients are essential in capturing the dynamics of the quadrotor accurately.

$$m \begin{bmatrix} \ddot{X} \\ \ddot{Y} \\ \ddot{Z} \end{bmatrix} = \begin{bmatrix} 0 \\ 0 \\ mg \end{bmatrix} - \begin{bmatrix} 0 \\ 0 \\ K_\tau \sum_i^4 \omega_i^2 \end{bmatrix} - \begin{bmatrix} K_{d_x} & 0 & 0 \\ 0 & K_{d_y} & 0 \\ 0 & 0 & K_{d_z} \end{bmatrix} \begin{bmatrix} \dot{X}^G \\ \dot{Y}^G \\ \dot{Z}^G \end{bmatrix} \quad (4)$$

In the literature, various methods have been proposed to estimate the drag [19] and thrust [20] coefficients. While numerical approximations are available, the most precise values are obtained through experimental tests. The following steps outline the procedure for conducting these experiments:

*1) Thrust Coefficient Effect*

The ratio between the square of the engine speed and the resulting thrust force which is also called as thrust coefficient ($K_T$), can be determined by equation of motion when quadrotor is in a hover position. In hover position the thrust force ($T$) can be written in terms of propeller diameter ($D$), linear velocity of the propellers, $\Delta v$ is the velocity of air accelerated by the propellers which is generally assumed as equal to $2v$, and $\rho$ is the air density that is assumed as $1.225 kg/m^3$ [21].

$$T = \frac{\pi}{4} D^2 \rho v \Delta v \quad (5)$$

Then the thrust coefficient can be found from:

$$T = K_T \omega^2 \quad (6)$$

By following these steps, the thrust force of the utilized quadrotor is computed as:

$$K_T = \frac{T}{\omega^2} = \frac{105.0588N}{2.0944e+03 rad/s} = 2.3950e-05 \quad (7)$$

*2) Drag Coefficient Effect*

The drag force, also referred to as air resistance, is generated when air particles are pushed by the propellers of the quadrotor. This force acts in the opposite direction to the UAV's movement and is influenced by factors such as air density, propeller characteristics, and engine rotation speed. Although there are operational differences between the thrust force and drag force, mathematically, the only distinction lies in the coefficients they are multiplied by.

To determine the drag coefficient, a test bench is typically required; however, analytical methods [15] and system identification techniques can also be utilized to converge or estimate this value. In our experiments, we estimated the drag coefficient as b=6.8429e-07.

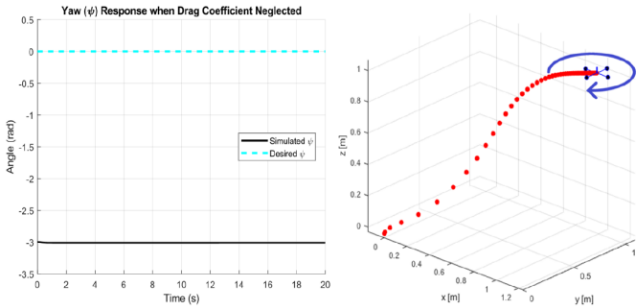

Fig. 1. Quadrotor Motion without drag coefficient

The drag and thrust coefficients play a crucial role in determining the dynamics of the quadrotor. Numerous publications suggest that the drag coefficient can be disregarded in calculations due to its relatively small value. In our experimental investigations, we observed that excluding the drag coefficient did not adversely affect the control of position, roll, and pitch. However, it is important to note that in the $U_4$ equation, the drag coefficient is utilized to determine the yaw moment. Consequently, neglecting this coefficient may result in continuous rotation around the yaw axis, even if the quadrotor achieves the desired position and maintains hover. The findings from our experiments conducted in this context corroborate this statement.

Once the translational and rotational dynamics, accounting for aerodynamic effects and motor torques, are derived, it becomes necessary to expand the calculations to include gyroscopic effects and the torque exerted on the roll, pitch, and yaw axes. This extension allows for a more comprehensive understanding of the quadrotor's behavior and dynamics. When all these computations solve in a line, the equation of the motion for a quadrotor can be obtained as:

$$\begin{bmatrix} \ddot{X}^G \\ \ddot{Y}^G \\ \ddot{Z}^G \\ \ddot{\phi} \\ \ddot{\theta} \\ \ddot{\psi} \end{bmatrix} = \begin{bmatrix} \frac{1}{m}[-(c\phi c\psi s\theta + s\phi s\psi)F_T^b - K_{d_x}\dot{X}^G] \\ \frac{1}{m}[-(c\phi s\psi s\theta - s\phi c\psi)F_T^b - K_{d_y}\dot{Y}^G] \\ \frac{1}{m}[-(c\phi c\theta)F_T^b - K_{d_z}\dot{Z}^G] + g \\ \frac{1}{J_x}[(J_y - J_z)qr - J_r q(\omega_1 - \omega_2 + \omega_3 - \omega_4) + lK_T(\omega_4^2 - \omega_2^2)] \\ \frac{1}{J_y}[(J_x - J_z)pr - J_r p(\omega_1 - \omega_2 + \omega_3 - \omega_4) + lK_T(\omega_1^2 - \omega_3^2)] \\ \frac{1}{J_z}[(J_z - J_y)pq - K_d(\omega_1^2 - \omega_2^2 + \omega_3^2 - \omega_4^2)] \end{bmatrix} \quad (8)$$

where $J_{x,y,z}$ is quadrotor body moment of inertia in all axes, $J_r$ is rotor moment of inertia, $l$ is length of the quadrotor arms, $g$ is gravitational acceleration, and $F_g, F_T^G$, and $F_T^b$ are gravity force, total thrust force in the global frame, and body frame respectively [22].

As it can be seen from the aforementioned steps, the quadrotor modeling is pretty challenging operation that needs many assumptions. Therefore, the obtained models do not always fully become compatible with actual systems.

*B. Controller Desing*

To control the grey-box and black-box systems in a simulated environment, linear controllers (PID and LQR) were designed based on the known model of the grey-box system. Experiments were conducted to construct both state space and transfer function models using system identification techniques, allowing estimation of unknown parameters within the grey-box system. The appropriate Q and R values for the LQR controller were determined using the state space model. Additionally, a multi-PID controller was applied to the transfer function model to compare the resulting positional outputs.

*1) LQR Controller Design*

System identification methods offer various approaches to estimate missing parameters of a system or controller based on the available data. In cases where only partial data is known, online estimation techniques like RLS (Recursive Least Squares) and Gradient Descent can be used. RLS, with its forgetting factor, provides better parameter estimation results, while Gradient Descent is more effective in reducing error to a desired level. Another model type, known as Model Reference Adaptive Control (MRAC), prioritizes performance over parameter estimation in tracking or regulation tasks. These approaches are useful for online parameter estimation without access to the full dataset. In the experiments conducted, state space matrices based on n4sid were obtained since all the data is available at this stage.

$$dx/dt = Ax(t) + Bu(t) + Ke(t) \quad (9)$$
$$y(t) = Cx(t) + Du(t) + e(t) \quad (10)$$

During the process, due to the complexity of analyzing 24 transfer functions between 4 inputs and 6 outputs, a grey-box model of the quadrotor is designed. System identification techniques are implemented to determine the state matrices of the grey-box model for the quadrotor. The LQR controller requires the Q and R parameters, along with the states. These parameters, known as performance matrices, determine the trade-off between energy consumption and system performance. The Q matrix, representing performance effort, is assigned a high value for critical system performance without concern for energy consumption. Alternatively, when energy consumption is equally important as system performance, the coefficients R and Q are chosen to be close to each other, typically between 0 and 1. To achieve the desired outcome, the Q and R parameters were adjusted multiple times until Q=1 and R=0.001 were determined. By utilizing the states and these parameter values, the feedback gain $K_f$ can be calculated using the relevant MATLAB command:

$$[K_f SE] = dlqry(A, B, C, D, Q, R) \quad (11)$$

Moreover, the reference gain $K_r$ can be computed by using the expression below:

$$K_r = (B^T S B + R)^{-1} B^T \left[ I - \left( A - B K_f \right)^T \right]^{-1} C^T Q \quad (12)$$

From these equalities, the reference ($K_r$) and feedback ($K_f$) gains are computed.

*2) PID Controller Design*

The PID control method minimizes deviation between desired and actual values of the plant. For the highly nonlinear quadrotor system, separate controllers are designed for position and attitude control. These controllers are independent but have interdependent parameters. To change the quadrotor's position, different torques are applied to the motors, which keeps the quadrotor oriented on the pitch and roll axes for movements in the forward-backward or right-left directions. This relationship between motion and attitude stabilization means that attitude control parameters affect the quadrotor's position. Therefore, it's crucial for position control that these parameters quickly reach desired values. Placing the position control block in the outer loop and the attitude and altitude control block in the inner loop enhances the control scheme, resulting in a more robust diagram. In the conducted experiments, a total of six PID controllers were employed to regulate both position and attitude errors. The equations derived for this purpose are as follows: The thrust force $U_1$ is determined based on the errors in the x, y, and z positions, whereas the attitude parameter errors are utilized to compute the control inputs $U_2$, $U_3$, and $U_4$.

$$U_1 = (g + K_{z,D}(z_d - z) + K_{z,P}(z_d - z))m/(C\phi C\theta) \quad (13)$$

$$U_2 = \left( K_{\phi,D}(\phi_d - \phi) + K_{\phi,P}(\phi_d - \phi) \right) J_{xx} \quad (14)$$

$$U_3 = \left( K_{\theta,D}(\theta_d - \theta) + K_{\theta,P}(\theta_d - \theta) \right) J_{yy} \quad (15)$$

$$U_4 = \left( K_{\psi,D}(\psi_d - \psi) + K_{\psi,P}(\psi_d - \psi) \right) J_{zz} \quad (16)$$

where $K_{axis,PID\,type}$ respresents the PID gains, and $\{z, \phi, \theta, \psi\}_d$ terms show the desired values. The controller for the x, y position comes from the error dynamics of outer loop:

$$\ddot{X} = (K_{x,P} + K_{x,I}/s)(x_d - x) \quad (17)$$

$$\ddot{Y} = (K_{y,P} + K_{y,I}/s)(y_d - y) \quad (18)$$

Based on the error dynamics, the PID controllers are implemented in the given equations and their gains are tuned using Simulink Auto Tuner. Finally, the LQR and PID controllers are applied to the grey-box nonlinear model of the quadrotor, and their performances are compared under the same input values. As it can be seen from the following figure the designed LQR controller outperformed the PID controller by considering the state space matrices of the model, which capture the characteristics of the nonlinear system. Consequently, the LQR controller is selected to compare the performance of the grey-box and black-box systems in the next section.

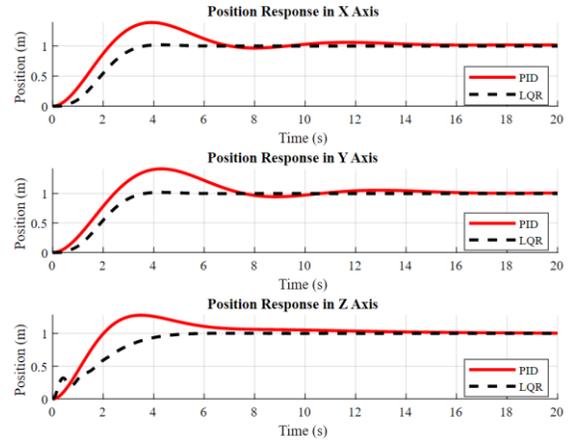

Fig. 2. Comparison of LQR and PID Controller on Position Control

*C. Modeling with System Identification*

System identification techniques play a crucial role in understanding, analyzing, and controlling dynamic systems, leading to enhanced system performance, fault detection, and informed decision-making, especially in situations where system details and models are not sufficiently clear. These techniques rely on input-output data to provide accurate models or estimate parameters that capture the system's characteristics. Therefore, the quality of the data used for system identification is crucial for consistent model accuracy.

In the conducted experiments, the quadrotor's position data was obtained from the onboard vision system, while the attitude data was obtained from the onboard IMU sensor. As for the choice of inputs, the angular velocities of the motors and the control inputs derived from angular velocities, as well as other numerical values such as motor arm length, thrust, and drag coefficients, were considered as possible options. By employing system identification techniques, the unknown or challenging-to-compute parameters like drag ($b$) and thrust coefficients ($K_T$) can be estimated, as demonstrated by the equations below.

$$U_1 = K_T(\omega_1^2 + \omega_2^2 + \omega_3^2 + \omega_4^2) \quad (19)$$

$$U_2 = lK_T(-\omega_2^2 + \omega_4^2) \quad (20)$$

$$U_3 = lK_T(-\omega_1^2 + \omega_3^2) \quad (21)$$

$$U_4 = b(-\omega_1^2 + \omega_2^2 - \omega_3^2 + \omega_4^2) \quad (22)$$

In this case, two scenarios were examined: when the input is selected as the angular velocities of the motors, and when the input is chosen as the control inputs with pre-known thrust and drag coefficients. The performance of the system responses was thoroughly evaluated, and the results clearly indicated that the estimated quadrotor model, utilizing angular velocities as inputs, yielded more consistent outcomes compared to the system model using control inputs. Therefore, the angular velocities of the motors are selected as input during the system identification process.

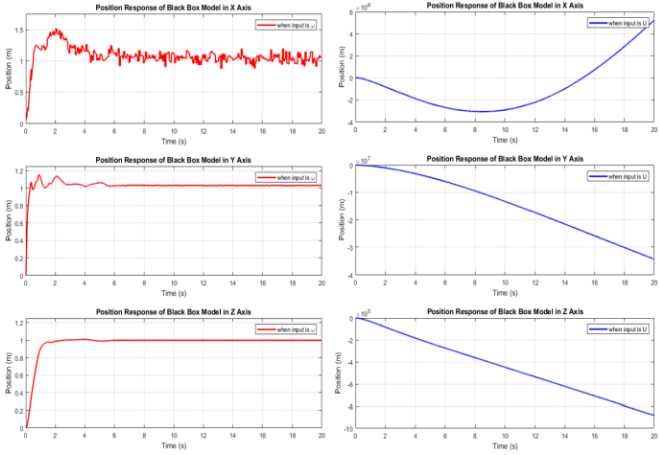

Fig. 3. Comparative Input Type Selection Effect on System Modeling

As mentioned in the previous section, where controller design was discussed, both transfer function and state space models of the grey-box system have been obtained and utilized for comparing controller performance. Considering the decision made in the previous section to evaluate simulation performance using the LQR controller, it has been determined that the state space model obtained from black-box system modeling process, is employed in the Simulink diagram.

In the scenario, the quadrotor hovers at a desired location one meter away in the x, y, and z directions. Data collection involves using a 3D-printed component and a mirror, enabling the quadrotor's onboard camera to monitor a ground marker. Attitude data is measured with the quadrotor's onboard IMU sensor. The quadrotor's motor speeds are recorded during motion towards the target position, and position and orientation data are obtained using the ArUco marker method and the onboard IMU sensor. The collected data, including position and attitude information, is then input into the system identification toolbox.

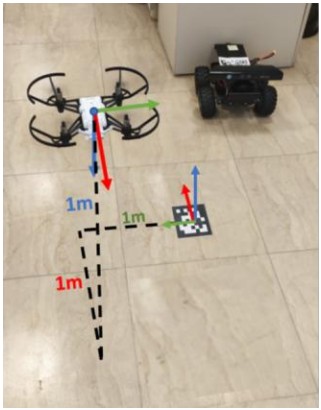

Fig. 4. During Flight Data Collection

In the experiments, attitude data was obtained from the IMU sensor, position data was collected using the onboard vision system with a ground marker, and instant motor velocities were recorded using the quadrotor's SDK. A total of 49,939 lines of input-output data were collected with a sampling time of 0.001 seconds over 50 seconds. The data were divided into 80% for estimation and 20% for validation during system identification. Both the black-box system estimated through system identification and the grey-box system designed in the simulation environment were subjected to identical inputs for performance comparison.

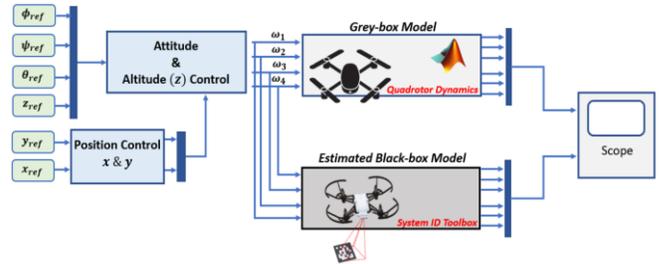

Fig. 5. Grey-box and Black-box Model Comparison with Same Input Signals and Same Controller

## III. RESULTS AND DISCUSSION

In this paper, grey-box and black-box system models were obtained using system identification techniques in MATLAB and Simulink. The LQR controller, which performed well in the grey-box model, was applied to control the black-box model derived from onboard vision system data. Both models were subjected to the same input and controller effects, and their responses were compared with experimental data.

The results show consistent performance among the black-box model, grey-box model, and experimental data from the onboard vision system, confirming the successful functioning of the camera and marker system. This instills confidence in the accuracy and reliability of the black-box model obtained through system identification. The consistent results further validate the effectiveness of system identification and affirm the functionality of the onboard camera and marker system.

The successful integration of the LQR controller with the black-box model highlights its compatibility and adaptability across different system models, showcasing its versatility and robustness as an effective control method. Overall, the comparative analysis between the grey-box and black-box models, along with the experimental data, demonstrates the reliability and effectiveness of the onboard vision system data in accurately representing the system dynamics.

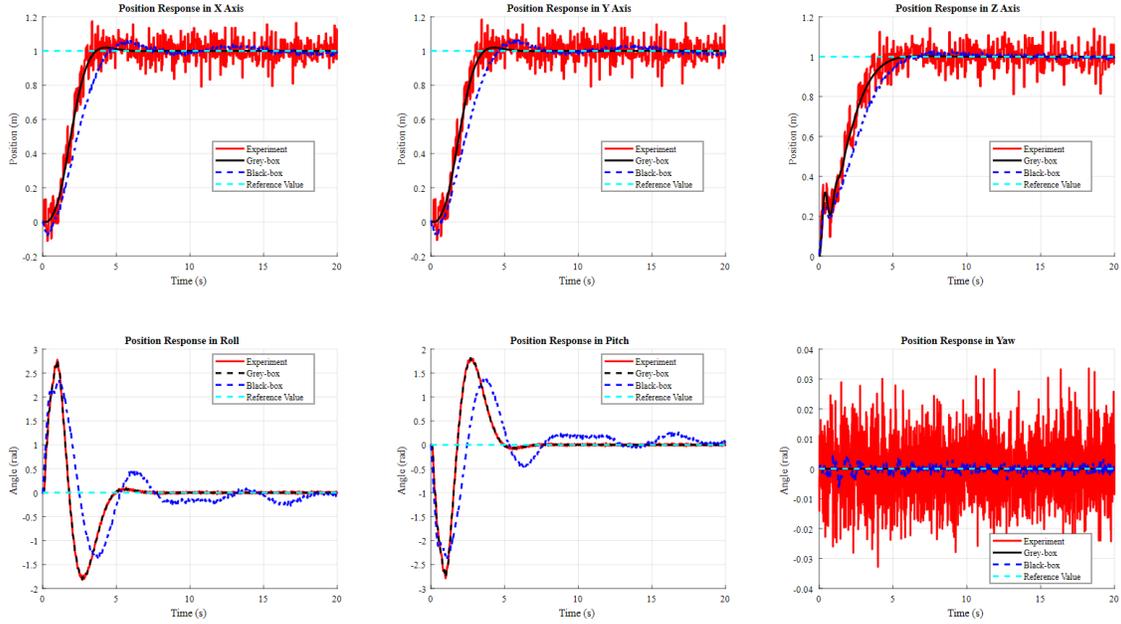

Fig. 6. Comparison of Position and Attitude Responses

## IV. Conclusion

In conclusion, this article explored the application of vision-based system identification techniques in quadrotor modeling and control. The challenges associated with quadrotor modeling were identified, including the complexity of numerical methods and the limitations caused by assumptions. Future research should focus on developing more efficient numerical methods and sophisticated modeling techniques to improve quadrotor modeling.

One of the main challenges in quadrotor modeling is the complexity of numerical methods. The motion equations of a quadrotor involve nonlinear dynamics and aerodynamic effects, which can be computationally intensive to solve accurately. Future studies can focus on developing more efficient and accurate numerical methods to improve quadrotor modeling. Another challenge lies in the assumptions made during quadrotor modeling. These assumptions simplify the model but may introduce limitations in representing the true dynamics of the system. Future research can explore more sophisticated modeling techniques that consider a broader range of factors, such as wind disturbances, payload variations, and nonlinearities, to improve the fidelity of quadrotor models.

Uncertainties in thrust and drag coefficients were addressed using grey-box modeling, and further research can refine this approach and explore additional uncertainty quantification techniques. Our study validated the effectiveness of vision-based system identification through the consistent performance of black-box and grey-box models using onboard vision system data. However, future studies should investigate alternative vision-based techniques, such as deep learning-based methods, for more accurate and robust quadrotor modeling and control.

Overall, our research demonstrates the potential of vision-based system identification techniques in enhancing quadrotor modeling and control. By integrating onboard vision system data, we achieved reliable representations of quadrotor dynamics, contributing to improved system modeling and advancing quadrotor performance. Future studies can further explore and utilize these techniques to enhance quadrotor performance, fault detection, and decision-making processes. Additionally, addressing challenges related to numerical complexity, model assumptions, and uncertainty quantification will be crucial for improving the accuracy and fidelity of quadrotor models.